\documentclass{article}
\usepackage{spconf,amsmath,graphicx}
\usepackage{booktabs}
\usepackage{colortbl}
\usepackage{algpseudocode}
\usepackage{algorithm} 
\usepackage{adjustbox}
\usepackage{color}
\usepackage{amsfonts}
\usepackage{amssymb}
\usepackage{bbding}
\usepackage{multirow}

\title{Deepening neural networks implicitly and locally via Recurrent Attention Strategy}
\name{Shanshan Zhong$^1$, Wushao Wen$^1$, Jinghui Qin$^{2\dagger}$\thanks{$^{\dagger}$Corresponding author. }, Zhongzhan Huang$^1$}
\address{$^1$ School of Computer Science and Engineering, Sun Yat-sen University\\
$^2$ School of Information Technology, Guangdong University of Technology
}

\begin{document}
%
\maketitle

\begin{abstract}
More and more empirical and theoretical evidence shows that deepening neural networks can effectively improve their performance under suitable training settings. However, deepening the backbone of neural networks will inevitably and significantly increase computation and parameter size. To mitigate these problems, we propose a simple-yet-effective Recurrent Attention Strategy (RAS), which implicitly increases the depth of neural networks with lightweight attention modules by local parameter sharing. The extensive experiments on three widely-used benchmark datasets demonstrate that RAS can improve the performance of neural networks at a slight addition of parameter size and computation, performing favorably against other existing well-known attention modules. 

\end{abstract}
\section{Introduction}
\label{sec:intro}

\begin{figure*}[t]
  \centering
  \includegraphics[width=0.98\linewidth]{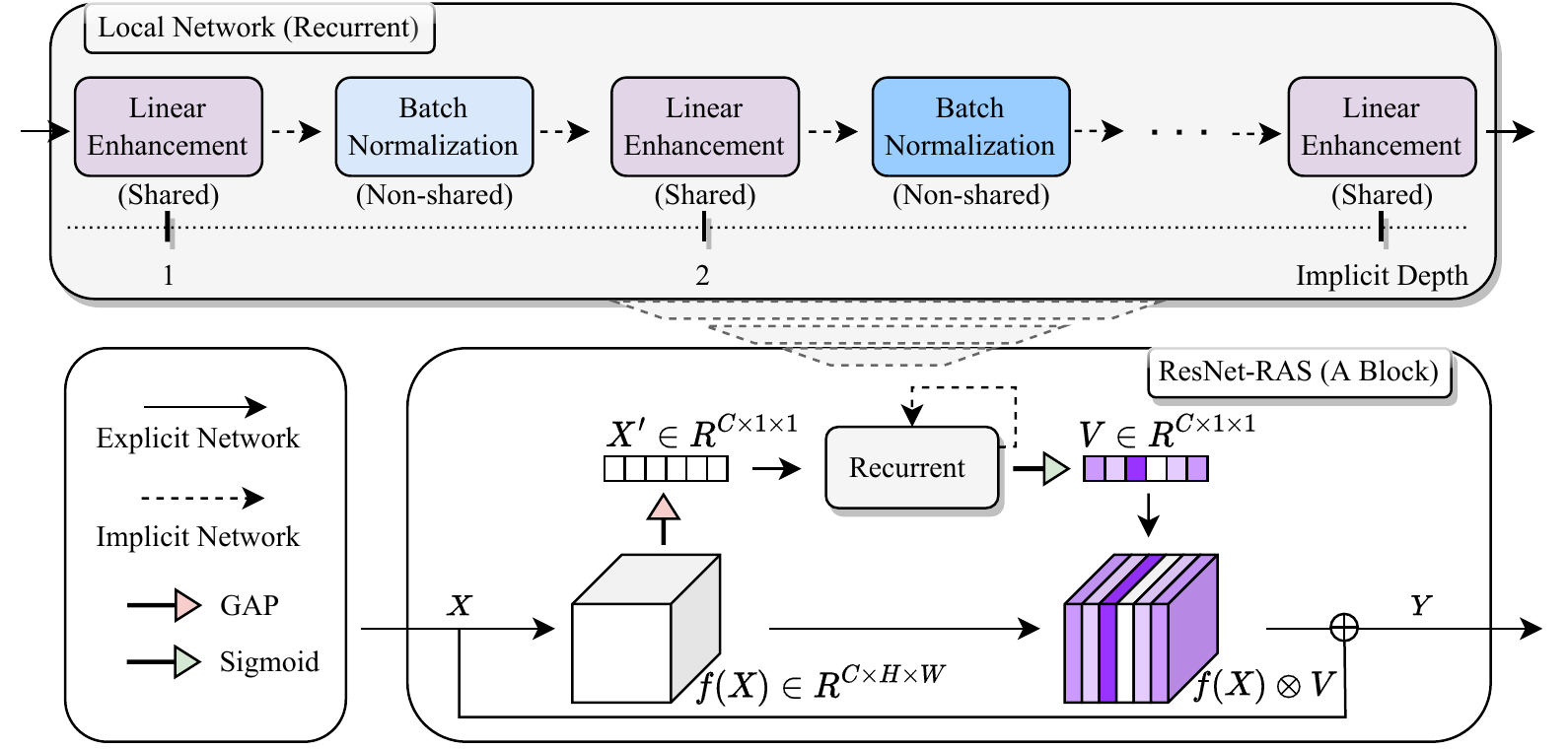}
  \caption{The residual block with RAS. $X$ is the input of the block, where $C, H, W$ represent its number of channels, height, and width respectively. $f(\cdot)$ is the residual operator, $g(\cdot)$ is the attention module, and $Y$ is the output of the residual block.}
\label{fig:ras}
\end{figure*}

More and more empirical and theoretical evidence~\cite{sun2016depth,he2016deep} shows that depth plays an important role in the success of Deep Neural Networks (DNNs). However, increasing the depth of DNNs by stacking network blocks inevitably and significantly increases parameter size (\#P) and computation, which is overwhelming in memory-bound industrial scenarios. We notice that attention mechanisms with simplicity and effectiveness have achieved great success in many visual tasks~\cite{guo2022attention,hassanin2022visual}. With the help of attention mechanisms~\cite{hu2018squeeze,huang2020dianet,woo2018cbam,huang2020efficient}, DNNs can process informative regions of images~\cite{anderson2005cognitive,zhang2021sa,hou2019weighted} more efficiently by paying attention to important information. 

In particular, we notice that embedding attention modules into the backbone can improve the performance of DNNs with less \#P than stacking more backbone blocks. 
Specifically, we consider ResNet~\cite{he2016deep} with popular attention method, i.e., Squeeze-and-Excitation module (SENet)~\cite{hu2018squeeze} as an example. ResNet83-SENet (94.23\%, 0.97M) can outperform ResNet164 (93.45\%, 1.7M) on CIFAR10 with less \#P. Moreover, ResNet83-SENet only needs extra 11.5\% \#P while ResNet164 is 2$\times$ larger than ResNet83 (93.16\%, 0.87M). 
This observation demonstrates that improving the performance of DNNs by embedding attention modules is more efficient than stacking more backbone modules. Therefore, different from the method achieving better performance by increasing the depth of backbone, we focus on another strategy: 

\textit{Can we only locally deepen attention modules instead of the backbone to efficiently achieve higher performance of DNNs?} 


To answer this question, we first take ResNet164-SENet as an example, whose forward processing for each block can be expressed as: 
\vspace{-2mm}
\begin{equation}
\begin{aligned}
Y = X + f(X) \otimes \sigma(g(X^{'})),
\end{aligned}
\label{eq:senet}
\end{equation}
where $X \in R^{C \times H \times W}$ is the input of the block, $Y \in R^{C \times H \times W}$ is the output, $f(\cdot)$ is the residual operator originated from the residual block, $\sigma$ is the Sigmoid function, $X^{'} = \mathbf{GAP}(f(X))$, and $g(\cdot)$ is the attention module. Generally, in SENet, $g(X^{'}) = W_1 (\delta(W_2 X^{'})),$ where $W_1$ and $W_2$ are the parameters of two full neural networks, and $\delta$ refers to the ReLU function. To explore how the depth of attention modules affects the performance, we stack deeper attention modules to explicitly increase the depth $k \in [0, \infty)$ of attention modules as:
\vspace{-2mm}
\begin{equation}
\begin{aligned}
g^k = W_1^{k} (\delta(W_2^{k} g^{k-1})),
\end{aligned}
\label{eq:senet-diff2}
\end{equation}
where $g^0 = X^{'}$. We set $k$ as 2 and find that the accuracy of ResNet164-SENet on CIFAR100 can be further improved by 0.43\%, which indicates that embedding deeper attention modules can achieve higher performance by locally increasing the depth of DNNs. However, explicitly increasing the depth of attention modules still requires additional \#P. To maintain the performance improvement without extra \#P and inspired by related works\cite{zhang2022minivit,shen2021sliced,wang2021recurrent,stelzer2021deep} which achieve model compression and parameter reduction by designing the recurrent strategy on the backbone, we regard $k$ as implicit depth and increase $k$ as follows:
\vspace{-2mm}
\begin{equation}
\begin{aligned}
g^k = W_1^{\text{share}} (\delta(W_2^{\text{share}} g^{k-1})),
\end{aligned}
\label{eq:senet-same}
\end{equation}
i.e., for any implicit depth $k$, we share the same learnable parameters $W_1^{\text{share}}$ and $W_2^{\text{share}}$. The experiment shows that this recurrent design can further increase the accuracy of ResNet164-SENet by 0.26\%. It motivates us to improve the performance of the backbone efficiently by increasing the depth of attention modules implicitly and locally. However, recurrence still makes computation multiplied although \#P is not increased. 

To save above problem, we propose a simple-yet-effective Recurrent Attention Strategy (RAS) which cyclically uses a lightweight attention module to ensure the extra computation is reasonable by taking full advantage of linear transformation with minimal \#P. 
We conduct extensive experiments on popular classification benchmarks, i.e., CIFAR10, CIFAR100, and STL10, showing that our RAS is more efficient while performing favorably against other existing well-known attention modules. Our contributions can be summarized as follows:
\vspace{-2mm}
\begin{itemize}
\item We propose a simple-yet-effective Recurrent Attention Strategy (RAS) to improve DNNs' performance by increasing depth implicitly and locally.
\item We further propose a lightweight attention mechanism to ensure the extra computation is reasonable with minimal \#P.
\item The extensive experiments on three widely-used datasets show that our RAS is more efficient while performing favorably against its counterparts.
\end{itemize}

\section{Recurrent Attention Strategy}
\label{sec:algorithm}
\begin{table*}[htbp]
  \centering
      \caption{Testing accuracy and Frames Per Second (FPS) on CIFAR10, CIFAR100, and STL10. ``\#P(M)" means the number of parameters (million). Bold and underline indicate the best results and the second best results, respectively. }
  \resizebox*{\linewidth}{!}{
        \begin{tabular}{lccccccccc}
    \toprule
    \multirow{2}[4]{*}{Model} & \multicolumn{3}{c}{CIFAR10} & \multicolumn{3}{c}{CIFAR100} & \multicolumn{3}{c}{STL10} \\
\cmidrule{2-10}          & \#P(M) & FPS & top-1 acc. (\%) & \#P(M) & FPS & top-1 acc. (\%) & \#P(M) & FPS & top-1 acc. (\%) \\
    \midrule
    ResNet164~\cite{he2016deep} & 1.70  & 4253 & 93.45 $\pm$ 0.20 & 1.73  & 4189 & 74.40 $\pm$ 0.34 & 1.70  & 559 & 83.78 $\pm$ 1.25 \\
    ResNet164-SENet~\cite{hu2018squeeze} & 1.91  & 3157  & 94.24 $\pm$ 0.20 & 1.93  & 3131  & 75.30 $\pm$ 0.36  & 1.91  & 520  & 84.76 $\pm$ 0.94  \\
    ResNet164-CBAM~\cite{woo2018cbam} & 1.92  &  1706 & 93.95 $\pm$ 0.11 & 1.94  & 1680  & 74.58 $\pm$ 0.20  & 1.92  & 241  & 84.01 $\pm$ 0.64  \\
    ResNet164-DIANet~\cite{huang2020dianet} & 1.92  & 2483  & 94.50 $\pm$ 0.15 & 1.95   & 2322 & \textbf{76.77 $\pm$ 0.12}  & 1.92  & 522  & 85.68 $\pm$ 0.36  \\
    ResNet164-ECA~\cite{2020ECA} & 1.70  & 3152  & 94.25 $\pm$ 0.12  & 1.73  & 3140  & 74.49 $\pm$ 0.37 & 1.70  & 521  & 83.99 $\pm$ 1.34  \\
    ResNet164-SEM~\cite{zhong2022switchable} & 1.95  & 2617  & \underline{94.79 $\pm$ 0.11} &  1.97  & 2409 & 76.59 $\pm$ 0.36 & 1.95  & 524  & \textbf{86.83 $\pm$ 0.06} \\
    ResNet164-RAS(Ours) & 1.74  &  3446 & \textbf{94.84 $\pm$ 0.14} & 1.76  &  3430 & \underline{76.74 $\pm$ 0.24} & 1.74  & 522  & \underline{86.15 $\pm$ 0.81} \\
    \midrule
    ResNet83~\cite{he2016deep} & 0.87  & 8076  & 93.16 $\pm$ 0.15      & 0.89   & 8028  & 73.55 $\pm$ 0.36       & 0.87   & 1066 & 82.04 $\pm$ 1.25\\
    ResNet83-SENet~\cite{hu2018squeeze} & 0.97  & 5950  & 94.23 $\pm$ 0.14  & 0.99  & 5925  & 74.91 $\pm$ 0.36  & 0.97  & 992  & 84.80 $\pm$ 0.92 \\
    ResNet83-CBAM~\cite{woo2018cbam} & 0.97    &  3286   &   93.31 $\pm$ 0.18     &   0.99  &   3198  &  73.24 $\pm$ 0.20   &  0.97   &  471   & 83.62 $\pm$ 0.66 \\
    ResNet83-DIANet~\cite{huang2020dianet} &  1.09   &  4688   & \underline{94.43 $\pm$ 0.09}       & 1.11   &   4637   &  \underline{75.62 $\pm$ 0.27}      &   1.09  &   984  & \underline{85.04 $\pm$ 0.36} \\
    ResNet83-ECA~\cite{2020ECA} & 0.87  &  6168 & 93.98 $\pm$ 0.36  & 0.89  & 5698  & 74.06 $\pm$ 0.36  & 0.87  &  991 & 81.34 $\pm$ 1.32 \\
    ResNet83-SEM~\cite{zhong2022switchable} & 0.99  & 4794  & 93.73 $\pm$ 0.17 & 1.01  & 4694   & 74.79 $\pm$ 0.08 & 0.99  & 997 & 84.50 $\pm$ 0.37 \\
    ResNet83-RAS(Ours) & 0.89  & 6545  & \textbf{94.47 $\pm$ 0.02} & 0.91  &  6515 & \textbf{75.85 $\pm$ 0.26} & 0.89  & 998  & \textbf{85.22 $\pm$ 0.20} \\
    \bottomrule
    \end{tabular}%
  }
  \label{tab:main}%
\end{table*}%


In Eq.(\ref{eq:senet-same}), we consider a simple recurrent strategy for SENet and achieve further performance improvement. In this section, we introduce a more efficient strategy, named Recurrent Attention Strategy (RAS) in Fig.\ref{fig:ras}, which can have significantly performance improvement with reasonable \#P and computation.

\subsection{Linear Enhancement} 

For an attention module such as SENet, although the computation of using this module once is not very large, the increase in computation is considerable if we forward a module recurrently. Therefore, we consider a lightweight-yet-efficient linear transformation~\cite{liang2020iebn} as $g(\cdot)$ shown in Eq.~(\ref{eq:ex}).
\vspace{-2mm}
\begin{equation}
\begin{aligned}
g(X^{'}) = X^{'} \times \gamma + \beta, 
\end{aligned}
\label{eq:ex}
\end{equation}
where two learnable parameters $\gamma$ and $\beta$ take charge of scaling and shifting the tensor $X^{'}$ to calibrate the representation power of feature map.  

\subsection{Implicit Deepening and Learnable Connection} 

As shown in Fig.~\ref{fig:ras}, we recurrently use the same attention module $g(\cdot)$ and 
connect the linear enhancement with individual Batch Normalization (BN).
We set $k$ be the implicit depth of attention modules, then the attention map $V$ is calculated as follows:
\vspace{-2mm}
\begin{equation}
\begin{aligned}
V = \sigma(g^k(X^{'}))= \sigma(g(\mathbf{BN}_{k-1}(g^{k-1}(X^{'}))),
\end{aligned}
\label{eq:recurrence}
\end{equation}
where $g^0 = X^{'}$ and $\mathbf{BN}_{k-1}$ is the batch normalization of $k$-th connection. After recurrence and $g(\cdot)$, we obtain the attention map $V$, and use $V$ to adjust the feature map like Eq.(\ref{eq:senet}), i.e., $Y = X + f(X) \otimes V$. In general, the value of $k$ is set to 2, which is discussed in Section \ref{sec:ablation}.

\section{Experiment}
\label{sec:experiment}

\subsection{Datasets and Implementation Details}

We evaluate RAS on STL10~\cite{coates2011analysis}, CIFAR10~\cite{krizhevsky2009learning} and CIFAR100~\cite{krizhevsky2009learning}. CIFAR10 and CIFAR100 have 50k train images and 10k test images of size 32 by 32 but have 10 and 100 classes respectively. STL10 has 5k train images and 8k test images of size 96 by 96 and has 10 classes. The batch size of STL10, CIFAR10, and CIFAR100 are 16, 128, and 128 respectively. 

We train all models with 
an Nvidia RTX 3080 GPU and set the epoch number to 164. SGD optimizer with a momentum of 0.9 and weight decay of $10^{-4}$ is applied. Furthermore, we use normalization and standard data augmentation, including random cropping and horizontal flipping during training. All experiments use ResNet based on bottleblock~\cite{he2016deep,huang2020dianet} as backbone.

\subsection{Image Classification}

We explore the effectiveness of RAS through image classification tasks. Table~\ref{tab:main} shows the results of several popular attention modules, including SENet~\cite{hu2018squeeze}, CBAM~\cite{woo2018cbam}, DIANet~\cite{huang2020dianet}, ECA~\cite{2020ECA} and SEM~\cite{zhong2022switchable} under ResNet38 and ResNet164. 
Specifically, with different training settings, we can find that RAS can achieve significant performance improvement and outperform most of the well-known attention modules.
When ResNet164 is used as the backbone, the performance of RAS is slightly lower than that of DIANet. Moreover, the number of RAS parameters is relatively small among the considered attention modules and increases by no more than 3\% for the backbone network. Although the number of additional parameters in ECA is minimal, RAS can significantly outperform the accuracy of ECA on CIFAR10, CIFAR100 and STL10 datasets.

In addition to performance and number of parameters, RAS achieves competitive results in terms of inference speed. Specifically, we use Frames Per Second (FPS) to measure the performance inference speed. On both CIFAR10 and CIFAR100 datasets, RAS can be significantly faster than other attention modules, including ECA with a smallest number of parameters. On the STL10 dataset, we can find that the inference speed of RAS is still satisfactory, despite the different attention modules have similar FPS.



\section{Ablation Study and Discussions}
\label{sec:ablation}
\subsection{Why we use $\mathbf{BN}$ as the connection of $g(\cdot)$?}


The activation function is a key component of building DNNs, which generally connects any adjacent blocks in the neural networks~\cite{he2016identity}. A neural network with implicit depth like RAS, also needs an activation function to connect the different feature maps generated by the attention module during the recurrent process. 

\begin{figure}[t]
  \centering
  \includegraphics[width=0.9\linewidth]{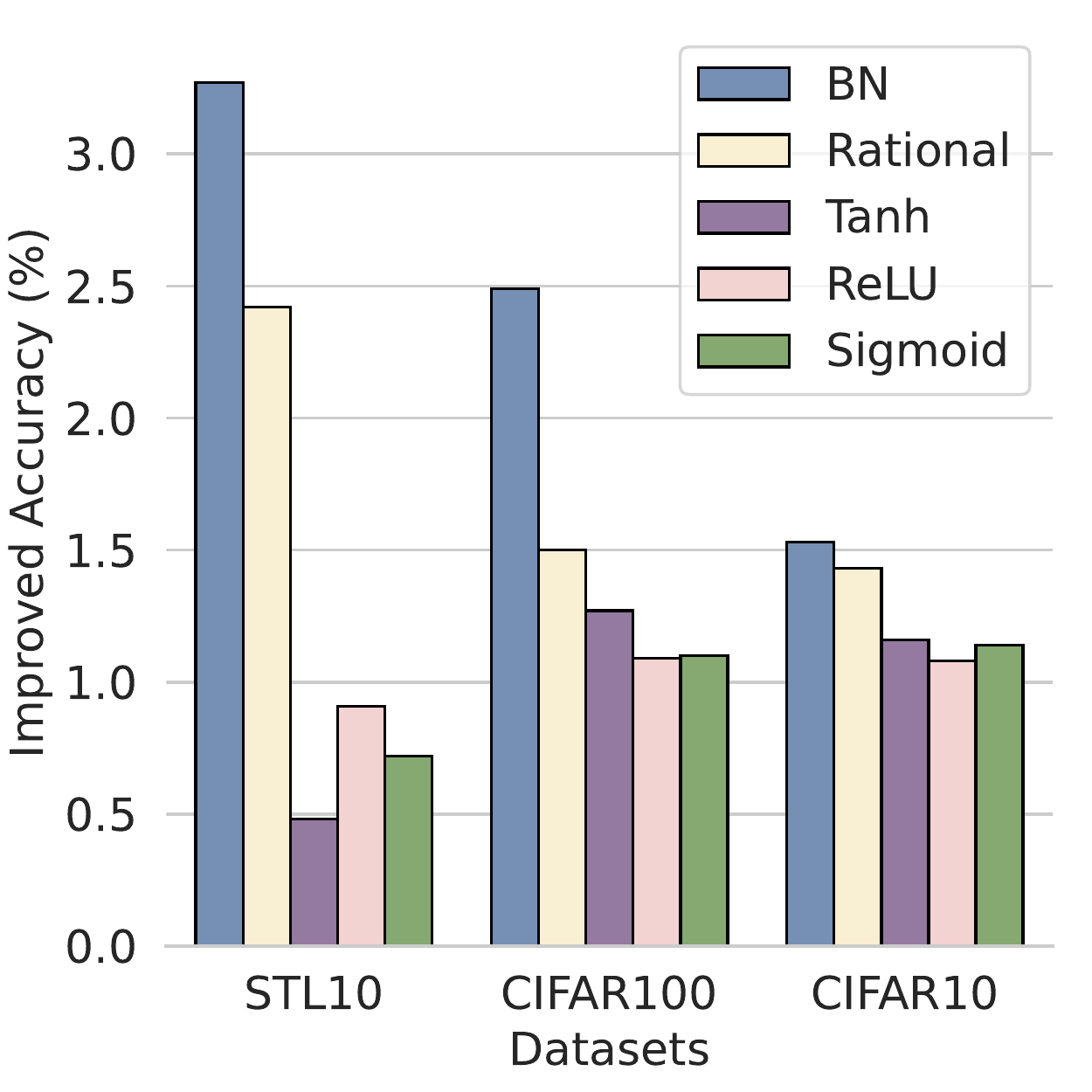}
  \caption{Accuracy(top-1 acc., \%) of ResNet164-RAS when different activation functions are used as the connection of recurrence on different datasets. }
\label{fig:connection}
\end{figure}

In previous works \cite{zhang2022minivit,shen2021sliced,wang2021recurrent,stelzer2021deep}, they considered the use of individual BN, which also is regarded as a kind of activation functions, to process the feature maps generated by the recurrent neural networks. Therefore, in Fig.~\ref{fig:connection}, we consider several activation function as candidates to connect the features generated by $g(\cdot)$.
Specifically, BN can obtain the best performance, which is consistent with the observation of these previous works. Although rational activation function \cite{molina2019pad}, which also has learnable parameters, can perform other activation functions, it still has a gap with BN. These results suggest that the features under RAS setting should be connected with linear activation rather than nonlinear activation.


\subsection{Why we set $k$ (implicit depth) as two?}


Under the RAS setting, the additional computational cost generated by the attention mechanism is positively related to $k$. If $k$ is large, it means that we have to forward the attention module many times, which will greatly reduce the efficiency of the model. Therefore we need to choose a suitable $k$ to obtain a trade-off of model speed and accuracy.


To explore the optimal $k$, We set the implicit depth from 1 to 3. As shown in Table~\ref{tab:depth}, when the depth exceeds 1, the performance of ResNet164-RAS decreases on different datasets. This is probably because attention modules are lightweight, which can effectively mine attention information in one cycle. 

Specifically, the experimental results recorded in Table~\ref{tab:depth} show that RAS with implicit depth of 2 is the most competitive. When depth is 1 which denotes that the number of cycles is 0, the performance of RAS on CIDAR10 and CIFAR100 is well, but the performance on STL10 only reaches 85.48\%, which is 0.67\% worse than the optimal depth, meaning complex large-scale data needs to deepen the depth of local networks to fully mine information. Besides, when the depth exceeds 2, RAS performs poorly on all datasets for the reason that the structure of linear enhancement is simple. Therefore, the recurrence of RAS reaches saturation when the depth is 2.




\begin{table}[htbp]
  \centering
  \caption{Accuracy(top-1 acc., \%) of ResNet164-RAS with different implicit depths on different datasets. }
  \resizebox*{0.85\linewidth}{!}{
    \begin{tabular}{cccc}
    \toprule
    Depth & \multicolumn{1}{l}{CIFAR10} & CIFAR100 & STL10 \\
    \midrule
    1     & 94.83 & 76.61 & 85.48 \\
    2     & 94.84 & 76.74 & 86.15 \\
    3     & 94.68 & 75.56 & 85.91 \\
    4     & 94.56 & 75.57 & 84.30 \\
    \bottomrule
    \end{tabular}%
  }
  \label{tab:depth}%
\end{table}%

\subsection{Setting of $\mathbf{BN}$}


As shown in Fig.~\ref{fig:ras}, Linear Enhancement is shared, but batch normalization is non-shared, in this section, we explore whether the batch normalization in RAS setting can be shared and how batch normalization affect the performance.

We test the performance of ResNet164 at depths 3 and 4 with attention modules on CIFAR100. From Table~\ref{tab:bn} we discover that non-shared BN is more stable among these attention modules. In addition, RAS is less affected by BN settings, indicating that RAS is a relatively stable strategy. It should be noted that attention modules show that the performance of depth 3 is stronger than that of depth 4, which is more obvious on the modules with larger parameters.
\begin{table}[htbp]
  \centering
  \caption{Accuracy(top-1 acc., \%) of ResNet164 with
different attention modules and implicit depth on CIFAR100. }
  \resizebox*{\linewidth}{!}{
    \begin{tabular}{lccccc}
    \toprule
    \multirow{2}[4]{*}{Module} & \multicolumn{1}{c}{\multirow{2}[4]{*}{\#P(M)}} & \multicolumn{2}{c}{Shared BN} & \multicolumn{2}{c}{Non-shared BN} \\
\cmidrule{3-6}          &       & Depth=3 & Depth=4 & Depth=3 & Depth=4 \\
    \midrule
    ECA\cite{2020ECA}   & 1.73  & 43.57 & 51.03 & 71.89 & 67.54 \\
    SENet~\cite{hu2018squeeze} & 1.93  & 72.19 & 70.16 & 75.47 & 74.74 \\
    SpaNet~\cite{guo2020spanet} & 3.86  & 71.38 & 70.55 & 74.37 & 69.19 \\
    RAS (Ours)   & 1.76  & 76.03 & 75.54 & 75.56 & 75.57 \\
    \bottomrule
    \end{tabular}%
    }
  \label{tab:bn}%
\end{table}%

\section{Conclusion}

In this paper, we discover that locally deepening attention modules can improve the performance of the backbone. To this end, we propose a Recurrent Attention Strategy to achieve an implicit deepening of the backbone through recurrent local networks. 
In particular, RAS designs linear transformation as attention modules, so that the performance of the backbone can be improved without significantly increasing the parameters and computation. Our experiments on three public datasets show that RAS outperforms multiple popular attention modules. Furthermore, ablation experiments demonstrate the rigorous design of RAS in detail.

\bibliographystyle{IEEEbib}
\bibliography{strings,refs}

\end{document}